\documentclass[letterpaper, 10 pt, conference]{ieeeconf}  

\IEEEoverridecommandlockouts                              

\usepackage{amsmath}
\usepackage{amssymb}

\usepackage{graphicx}
\usepackage{subcaption}
\usepackage[table]{xcolor}
\usepackage{booktabs}
\usepackage{adjustbox}
\usepackage{multirow}

\usepackage{algorithm}
\usepackage{algpseudocode}

\usepackage{url}

\usepackage{cite}

\usepackage{hyperref}

\title{\LARGE \bf
Learning Safe Autonomous Driving Policies Using Predictive Safety Representations
}

\author{Mahesh Keswani$^{1}$ and Raunak Bhattacharyya$^{1}$%
\thanks{$^{1}$Yardi School of Artificial Intelligence, 
        Indian Institute of Technology Delhi
        {\tt\small \{aiy247544, raunakbh\}@iitd.ac.in}}%
\thanks{This work has been submitted to the IEEE for possible publication. Copyright may be transferred without notice, after which this version may no longer be accessible.}
}

\begin{document}

\maketitle
\thispagestyle{empty}
\pagestyle{empty}

\begin{abstract}
Safe reinforcement learning (SafeRL) is a prominent paradigm for autonomous driving, where agents are required to optimize performance under strict safety requirements. This dual objective creates a fundamental tension, as overly conservative policies limit driving efficiency while aggressive exploration risks safety violations. The Safety Representations for Safer Policy Learning (SRPL) framework addresses this challenge by equipping agents with a predictive model of future constraint violations and has shown promise in controlled environments. This paper investigates whether SRPL extends to real-world autonomous driving scenarios. Systematic experiments on the Waymo Open Motion Dataset (WOMD) and NuPlan demonstrate that SRPL can improve the reward–safety tradeoff, achieving statistically significant improvements in success rate (effect sizes $r = 0.65-0.86$) and cost reduction (effect sizes $r = 0.70-0.83$), with $p < 0.05$ for observed improvements. However, its effectiveness depends on the underlying policy optimizer and the dataset distribution. The results further show that predictive safety representations play a critical role in improving robustness to observation noise. Additionally, in zero-shot cross-dataset evaluation, SRPL-augmented agents demonstrate improved generalization compared to non-SRPL methods. These findings collectively demonstrate the potential of predictive safety representations to strengthen SafeRL for autonomous driving.
\end{abstract}

\section{INTRODUCTION}

The development of autonomous driving technology presents an opportunity to improve transportation safety and efficiency by mitigating risks associated with human error~\cite{who_road_traffic_2023}. While traditional rule-based control systems have been foundational, their inherent brittleness struggles to address the complexity of real-world driving scenarios \cite{aksjonov2021rule}. This limitation has motivated a shift towards learning-based approaches, with reinforcement learning emerging as a promising paradigm. Reinforcement learning (RL) enables agents to learn optimal driving policies and adapt to novel situations through experience rather than manual programming~\cite{9351818}. 

Although deep RL has shown considerable success in continuous control tasks~\cite{duan2016benchmarking, fujimoto2018td3}, standard RL formulations optimize for reward maximization without explicit safety constraints, which can lead to dangerous behaviors in safety-critical applications~\cite{wachi2023safe}. Safe reinforcement learning (SafeRL) addresses this by formulating tasks as Constrained Markov Decision Processes (CMDPs), seeking policies that optimize performance while satisfying constraints on undesirable outcomes such as collisions~\cite{altman2021constrained}.

\begin{figure}[t]
    \centering
    \includegraphics[width=\linewidth]{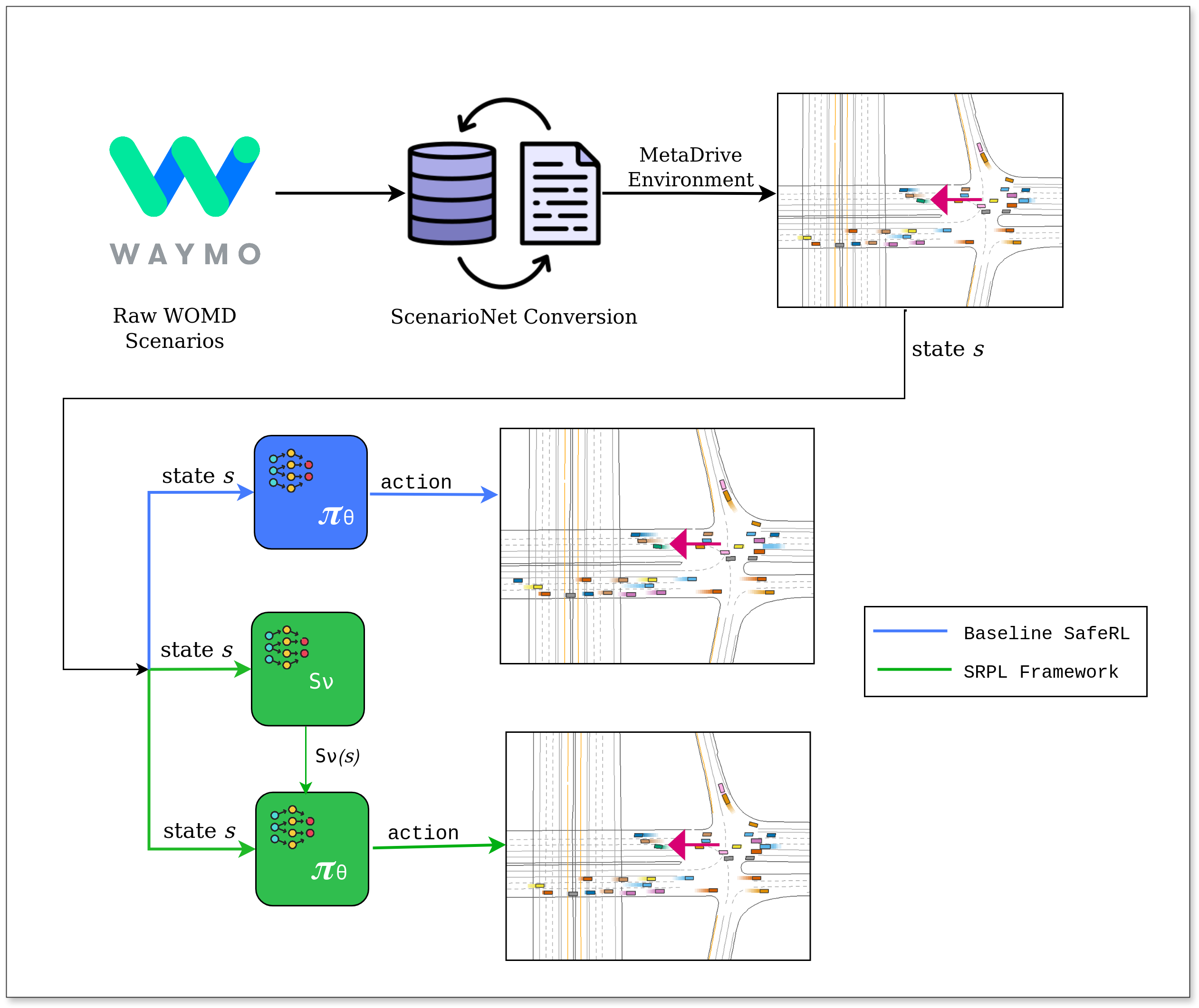}
    \caption{Integration of SRPL with SafeRL algorithms: Raw Waymo Open Motion Dataset (WOMD) scenarios are converted to MetaDrive-compatible format using ScenarioNet, generating bird's-eye view driving environments. The baseline SafeRL approach (blue) uses raw state observations as direct input to policy $\pi_\theta$. The SRPL framework (green) augments decision-making by concatenating raw states with predictive safety information from the Steps-to-Cost model $S_\nu(s)$ before policy execution. Pink arrows indicate the ego-agent's position in each scenario.}
    \label{fig:srpl_flow}
\end{figure}

Despite its theoretical advantages, the practical application of SafeRL in autonomous driving has revealed significant challenges. A recognized issue is the tendency of SafeRL agents to develop conservative behaviors, a phenomenon partly attributed to primacy bias, where initial encounters with trajectories violating constraints can persistently hinder agents' willingness to explore their environment \cite{nikishin2022primacy}. To encourage safety during training, SafeRL algorithms employ approaches such as failure penalties and explicit safety constraints during exploration~\cite{xie2022ask, achiam2017constrained}. Although these methods effectively minimize unsafe behaviors, they often lead to conservative policies that sacrifice performance for constraint satisfaction~\cite{mani2025safety}. Beyond the exploration-safety balance, real-world autonomous driving deployment faces additional challenges such as sensor noise and diverse traffic patterns that differ from controlled training environments. These factors can reduce policy performance and compromise the safety guarantees established during training. Therefore, the primary difficulty in applying SafeRL in autonomous driving is developing agents that can achieve high task performance without violating safety constraints while remaining robust to environmental variations.

To address these challenges, an alternative approach is to augment agents' decision-making capabilities with predictive models that anticipate future constraint violations. This enables agents to make better-informed decisions by anticipating potential constraint violations during exploration. The Safety Representations for Safer Policy Learning (SRPL) framework \cite{mani2025safety} implements this concept by augmenting the state with information from a learned Steps-to-Cost (S2C) model, which provides a probabilistic distribution of potential constraint violations over a future horizon. By incorporating these predictive safety representations, agents can make more nuanced decisions about when to explore and exercise caution.

Investigating the real-world viability of the SRPL framework in autonomous driving, therefore, requires a comprehensive evaluation under conditions that mimic real-world deployment challenges such as sensor noise and domain variations. Such a comprehensive evaluation includes not only measuring standard performance metrics, but also systematically assessing the robustness of learned policies to sensor noise and their ability to generalize across diverse driving domains.

In this paper, we make the following contributions:
\begin{itemize}
    \item We provide a systematic evaluation of SRPL's effectiveness in real-world autonomous driving scenarios, demonstrating statistically significant improvements in success rates and cost reduction across multiple SafeRL algorithms on Waymo Open Motion Dataset (WOMD) and NuPlan datasets, while revealing algorithm-specific and context-dependent performance patterns.

    \item We show through robustness analysis that these safety representations improve resilience to observational noise and provide insights into the policy output stabilization effect underlying this robustness benefit.
 
    \item We identify asymmetric cross-dataset transfer, where agents trained on the diverse WOMD dataset generalize better to NuPlan than the reverse. In addition, we empirically demonstrate that SRPL augmentation improves domain generalization compared to non-SRPL methods.
    
    \item We establish practical guidance for SafeRL algorithm selection through comprehensive cross-dataset and robustness evaluations, identifying which algorithms benefit from SRPL augmentation under different deployment scenarios.
\end{itemize}

\section{RELATED WORK}

The field of learning-based autonomous driving has recently shifted its focus from procedurally generated scenarios \cite{chowdhury2023graphbasedpredictionplanningpolicy, chowdhury2024deepattentiondrivenreinforcement} to large-scale, real-world datasets like WOMD \cite{ettinger2021large} and NuPlan \cite{caesar2021nuplan}. While these datasets have been valuable for motion forecasting and planning \cite{shi2023motiontransformerglobalintention, seff2023motionlm, caesar2021nuplan}, decision-making and control in autonomous driving remain challenging.

Representation learning has become crucial to address these decision-making and control challenges, where compact representations capturing task-relevant information improve performance and sample efficiency. This is often achieved using auxiliary training signals or predicting future latent states \cite{jaderberg2016reinforcementlearningunsupervisedauxiliary, NIPS2019_8724, abs-1803-10122}. A promising direction within this area is to develop representations that explicitly encode safety-relevant information by learning to predict proximity to future constraint violation, so agents can make more informed decisions \cite{pmlr-v119-ota20a, schwarzer2021dataefficient, mani2025safety}.

Control policies for autonomous driving can be learned through two main paradigms: Imitation Learning (IL), which is sample-efficient but struggles with out-of-distribution scenarios, and RL, which discovers more robust policies but faces sample efficiency and safety challenges \cite{lu2023imitation}. While recent work explores hybrid IL-RL methods \cite{lu2023imitation, booher2024cimrl}, this paper focuses specifically on SafeRL algorithms augmented with predictive safety representations to understand their isolated effectiveness in real-world scenarios.

Concurrent work by \cite{charraut2025vmax} introduces V-Max, a comprehensive RL framework built on Waymax that provides standardized tools for training and evaluating various standard RL and IL algorithms in autonomous driving scenarios. While V-Max focuses on benchmarking different algorithmic approaches and system components (observation shaping, network architectures, reward shaping), our work specifically investigates how augmenting existing SafeRL algorithms with predictive safety representations affects their performance, robustness, and cross-domain generalization.

\section{PRELIMINARIES}

\subsection{Markov Decision Processes (MDPs)}
A Markov Decision Process (MDP) \cite{sutton1998reinforcement} is defined by a tuple $\langle \mathcal{S}, \mathcal{A}, \mathcal{P}, \mathcal{R}, \mu_0, \gamma \rangle$, where $\mathcal{S}$ is a set of states, $\mathcal{A}$ is a set of actions, $\mathcal{R}: \mathcal{S} \times \mathcal{A} \rightarrow \mathbb{R}$ is the reward function, $\mathcal{P}: \mathcal{S} \times \mathcal{A} \times \mathcal{S} \rightarrow [0,1]$ is the state transition probability function, $\mu_0: \mathcal{S} \rightarrow [0,1]$ is the initial state distribution, and $\gamma \in [0,1)$ is the discount factor. A policy $\pi_\theta$, parameterized by $\theta$, maps states to a probability distribution over actions. The goal in an MDP is to find an optimal policy $\pi_\theta^*$ that maximizes the expected discounted cumulative reward $J^R(\pi_\theta) = \mathbb{E}_{\tau \sim p_{\pi_\theta}(\tau)} \left[ \sum_{t=0}^{\infty} \gamma^t R(s_t, a_t) \right]$, where $\tau$ represents a complete state-action trajectory sampled by following policy $\pi_\theta$.

\subsection{Constrained Markov Decision Processes (CMDPs)}
To address the critical safety requirements of autonomous driving, the standard MDP is extended to a Constrained Markov Decision Process (CMDP) \cite{altman2021constrained}, which introduces a cost function $\mathcal{C}: \mathcal{S} \times \mathcal{A} \rightarrow \mathbb{R}^+$ and a corresponding safety threshold $\kappa$. The cost function quantifies undesirable outcomes, such as collisions. The objective in a CMDP is to find an optimal policy $\pi_\theta^* = \arg\max_{\pi_\theta} J^{R}(\pi_\theta)$ subject to $J^{C}(\pi_\theta) \leq \kappa$, where the expected cost $J^{C}(\pi_\theta)$ is defined analogously to the expected reward.

\section{METHODOLOGY}

\begin{algorithm}
\caption{SRPL for Autonomous Driving}
\label{alg:srpl_driving}
\begin{algorithmic}[1]
\State \textbf{Input:} Autonomous driving environment, safety horizon $H_s$, bin size $b$
\State \textbf{Initialize:} Driving policy $\pi_\theta$, S2C model $S_{\nu}$, target S2C model $S_{\nu'}$, S2C buffer $B_s$

\For{each driving scenario}
    \State Collect trajectory $\tau = (s_0, a_0, c_0, r_0, s_1, \dots)$.
    
    \State \textit{// Label states with steps-to-cost}
    \State $next\_violation\_idx \leftarrow \infty$
    \For{$t = |\tau|-1$ \textbf{to} $0$}
        \If{$c_t > 0$}
            \State $steps\_to\_violation_t \leftarrow 0$
            \State $next\_violation\_idx \leftarrow t$
        \Else
            \State $steps\_to\_violation_t \leftarrow \min(next\_violation\_idx - t, H_s)$
        \EndIf
        
        \State \textit{// Discretize into bins}
        \State $binned\_steps\_to\_violation_t \leftarrow \min(\lfloor steps\_to\_violation_t / b \rfloor, \lfloor H_s / b \rfloor - 1)$
        \State Add $(s_t, binned\_steps\_to\_violation_t)$ to $B_s$.
    \EndFor
    
    \State \textit{// Update S2C model}
    \State Sample a mini-batch $(s_j, binned\_h_j)$ from $B_s$.
    \State Update $\nu$ of $S_{\nu}$ by minimizing the loss in Eq. \eqref{eq:s2c_loss}.
    \State Periodically update target model: $S_{\nu'} \leftarrow S_{\nu}$.
    
    \State \textit{// Update policy with augmented state}
    \For{each policy update step}
        \State $s_{aug} \leftarrow [s \oplus S_{\nu'}(s)]$
        \State Update $\pi_\theta$ using a SafeRL algorithm with $s_{aug}$.
    \EndFor
\EndFor
\State \textbf{Return:} Trained policy $\pi_\theta$, S2C model $S_{\nu}$
\end{algorithmic}
\end{algorithm}

This section details the core components of our methodological approach. We first introduce the SRPL framework and then define the autonomous driving task formulation.

\subsection{Safety Representations for Safer Policy Learning (SRPL)}
The SRPL framework equips agents with a predictive model of future constraint violations \cite{mani2025safety}. The core of the approach, detailed in Algo.~\ref{alg:srpl_driving}, is a Steps-to-Cost (S2C) model, \(S_\nu : \mathcal{S} \rightarrow \Delta H_s\), which is a probabilistic model that, for any given state \(s\), predicts the distribution over timesteps until a potential constraint violation occurs:
\[
S_\nu^t (s) = P(\delta(s) = t | s)
\]
Here, \(\delta(s)\) is the "steps-to-cost" value, a random variable representing the number of steps from the current state until a constraint violation occurs. The prediction is made over a fixed lookahead window called the Safety Horizon (\(H_s\)). To make this prediction tractable, SRPL discretizes this horizon into a set of categorical bins. The S2C model is framed as a classification problem, where it learns to output a probability distribution over these discrete bins.

To learn this predictive distribution, the S2C model's supervised learning update is interleaved with the reinforcement learning process. Specifically, its parameters \(\nu\) are updated periodically by minimizing the negative log-likelihood loss over a dedicated safety buffer $B_s$:
\begin{equation}
\mathcal{L}(\nu) = -\mathbb{E}_{(s,y) \sim \mathcal{B}_{\text{s}}} [\log S_\nu(y|s)]
\label{eq:s2c_loss}
\end{equation}

where \(y\) is the ground-truth bin index corresponding to the number of timesteps to constraint violation that is observed from state $s$ during a rollout. This interleaved training ensures that the S2C model continuously adapts to the evolving state-visitation distribution of the policy. The S2C model integrates with the policy network through simple state concatenation, a process illustrated in Fig.~\ref{fig:srpl_flow}:
\[
\pi_\theta : [s \oplus S_\nu(s)] \rightarrow \mathcal{P}(\mathcal{A})
\]
where \(\mathcal{P}(\mathcal{A})\) represents a probability distribution over the action space. The dedicated safety buffer allows the S2C model to learn efficiently from the entire history of safety-relevant experiences, improving data efficiency for safety representation learning. In addition, the S2C model is implemented as a lightweight feedforward network, with inference averaging 52 milliseconds per forward pass, representing minimal computational overhead for real-time autonomous driving applications.

\subsection{Task Setup}

Our autonomous driving task formulation is adapted from ScenarioNet \cite{li2023scenarionet}, providing a standardized framework for real-world scenario evaluation. The task consists of four key components: observation space, action space, reward function, and cost function.

\textbf{Observation.} The agent's observation is a vector composed of three primary components: environmental perception, vehicle state, and navigation guidance. Environmental perception is provided by a 120-dimensional Lidar-like vector for detecting obstacles and a 12-dimensional vector for identifying drivable area boundaries, both with a 50-meter range. The ego-vehicle's state is summarized by its current steering angle, heading, velocity, and relative distance to the reference path. All sensor inputs are normalized to the [0, 1] range. Finally, navigation guidance is provided as a sequence of 10 future waypoints along the planned trajectory, projected into the vehicle's local coordinate frame to direct it toward its destination.

\textbf{Action.} The driving policy operates end-to-end, directly mapping sensor observations to low-level vehicle controls. The policy outputs a continuous, two-dimensional action vector for acceleration and steering, with each component normalized to the range [-1, 1]. These normalized values are then scaled to correspond to the vehicle's physical acceleration and steering commands.

\textbf{Reward Function.} The reward function \(R(s_t, a_t)\) is composed of both dense, per-timestep components and a sparse terminal reward. The dense rewards encourage driving progress, correct heading, and lane adherence. Formally, at each timestep \(t\), the dense portion of the reward is given by:
\[
R(s_t, a_t) = w_{\text{drive}} \cdot r_{\text{progress}} - w_{\text{heading}} \cdot p_{\text{heading}} - w_{\text{lat}} \cdot p_{\text{lateral}}
\]
The \(r_{\text{progress}}\) term denotes the longitudinal distance the vehicle travels along the reference route between two consecutive timesteps, providing a dense reward to encourage forward movement. The \(p_{\text{heading}}\) term is a penalty proportional to the heading error, discouraging deviation from the route's direction. Finally, \(p_{\text{lateral}}\) is a penalty for the lateral distance from the lane centerline, up to a maximum of 2 meters. The corresponding weights are set to \(w_{\text{drive}} = 1\), \(w_{\text{heading}} = 1\), and \(w_{\text{lat}} = 1\).

In addition, a sparse terminal reward, \(r_{\text{terminal}}\), is awarded at the end to incentivize task completion. A reward of +10 is given if the agent successfully reaches its destination, and a penalty of -5 is applied if the episode terminates due to other reasons, such as timing out far from the goal.

\textbf{Cost Function.} The cost function \(C(s_t, a_t)\) exclusively penalizes safety-critical events. An agent incurs costs if it collides with another entity or drives off the road. The cost function is defined as:
\[
C(s_t, a_t) = w_{\text{crash}} \cdot \mathbb{I}_{\text{crash}} + w_{\text{oor}} \cdot \mathbb{I}_{\text{out\_of\_road}}
\]
Here, \(\mathbb{I}_{\text{crash}}\) is an indicator function that equals one if a collision with a vehicle or pedestrian occurs, and zero otherwise. Similarly, \(\mathbb{I}_{\text{out\_of\_road}}\) is an indicator function that becomes one if the agent's vehicle leaves the drivable area. Both penalty weights, \(w_{\text{crash}}\) and \(w_{\text{oor}}\), are set to 2.

\section{EVALUATION RESULTS}
\setlength{\abovecaptionskip}{\baselineskip}
\setlength{\belowcaptionskip}{\baselineskip}
\begin{figure}[t]
\centering

\setlength{\abovecaptionskip}{5pt}
\setlength{\belowcaptionskip}{0pt}

\begin{subfigure}{0.49\textwidth}
    \includegraphics[width=\linewidth]{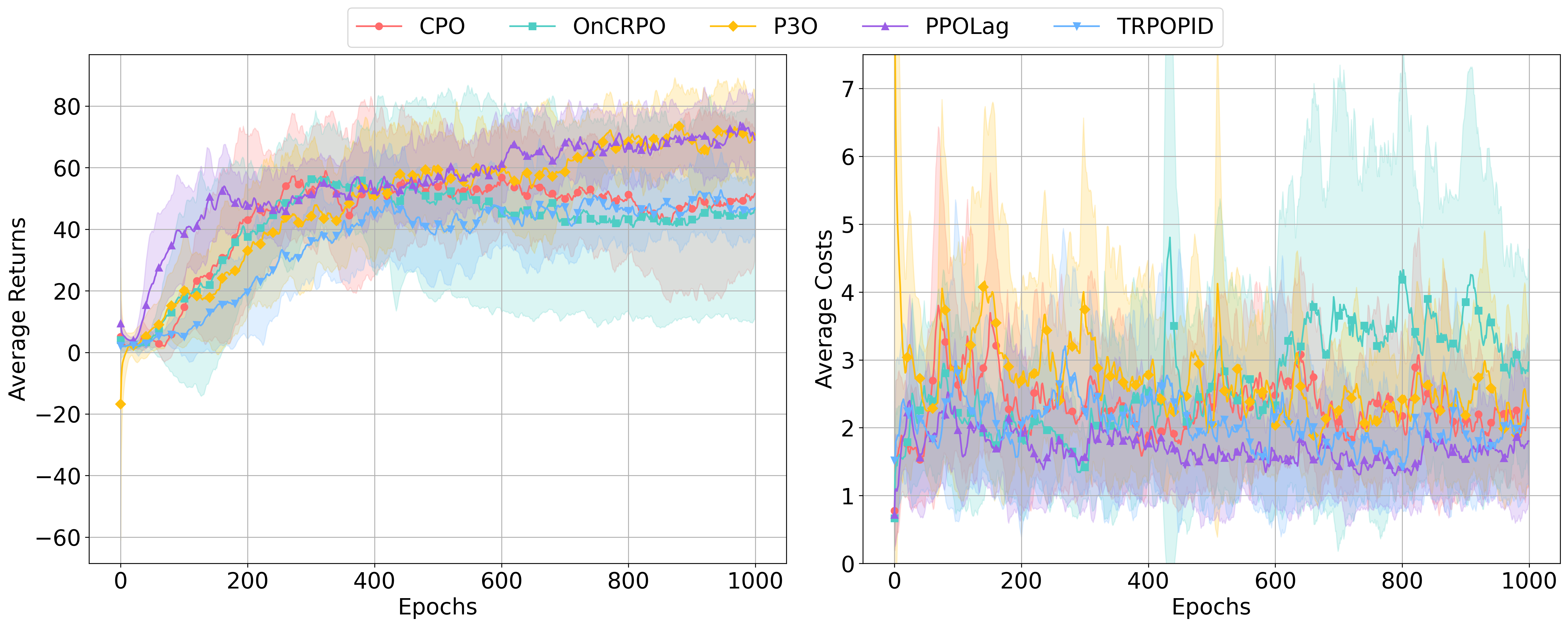}
    \caption{Training curves for baseline SafeRL algorithms.}
    \label{fig:baseline_combined}
\end{subfigure}
\hfill 
\begin{subfigure}{0.49\textwidth}
    \includegraphics[width=\linewidth]{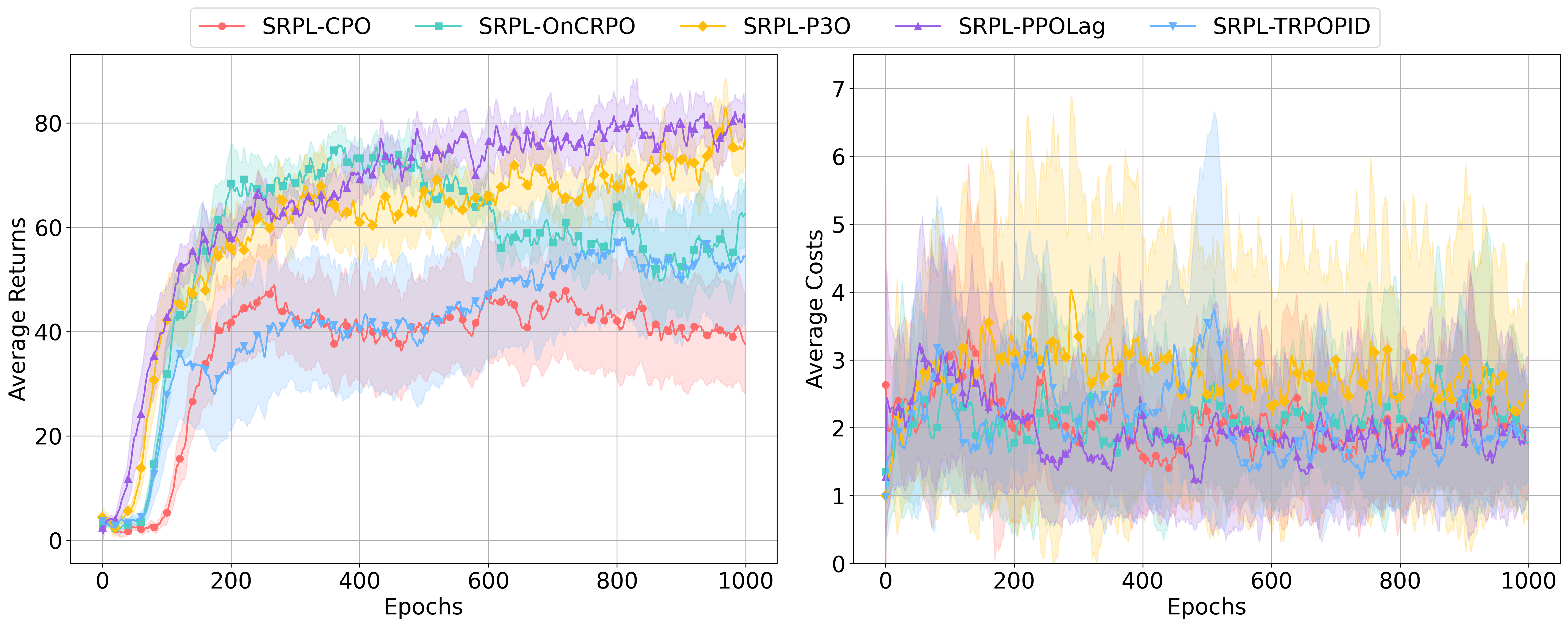}
    \caption{Training curves for corresponding SRPL-augmented algorithms.}
    \label{fig:srpl_combined}
\end{subfigure}

\caption{Comparison of training performance on WOMD between baseline SafeRL algorithms and their SRPL-augmented counterparts. The left plots show average returns (higher is better), while the right plots show average costs (lower is better).}
\label{fig:training_curves}
\end{figure}

\subsection{Experiment Setting}

\subsubsection{Scenario Datasets and Simulation}
\label{subsubsec:exp_setting}
Our experiments are conducted on two driving datasets: WOMD \cite{ettinger2021large} and NuPlan \cite{caesar2021nuplan}. For both datasets, we sampled a training set of 5,000 scenarios and a separate validation set of 1,000 scenarios, with NuPlan scenarios drawn from the Boston region. To ensure the fidelity of our analysis, we computed key complexity metrics from our sampled subsets and compared them against published statistics from the complete datasets. For instance, the sampled WOMD training set contains an average of 85.18 vehicles, 130.45 meters track length, an intersection ratio of 0.69, and 11.90 pedestrians per scenario, closely matching the full-dataset statistics \cite{li2023scenarionet}. Similarly, the sampled NuPlan scenarios contain an average of 52.18 vehicles, a track length of 90.42 meters, an intersection ratio of 0.54, and 22.06 pedestrians per scenario, which aligns with the statistics of the NuPlan Boston dataset \cite{li2023scenarionet}.

To enable interactive learning, we used the MetaDrive simulator \cite{li2022metadrive} to replay the scenarios from both WOMD and NuPlan. This provides a consistent simulation environment with reactive background agents, allowing the ego-vehicle to interact dynamically with the world.

Evaluation includes five prominent on-policy SafeRL algorithms: Constrained Policy Optimization (CPO) \cite{achiam2017constrained}, PPO-Lagrangian (PPOLag) \cite{ray2019benchmarking}, Trust Region Policy Optimization with PID (TRPOPID) \cite{stooke2020responsive}, Online Constrained Rectified Policy Optimization (OnCRPO) \cite{xu2021crpo}, and Penalized Proximal Policy Optimization (P3O) \cite{zhang2022penalized}. These baselines were selected because they represent diverse strategies for handling constraints in SafeRL.

\setlength{\abovecaptionskip}{\baselineskip}
\setlength{\belowcaptionskip}{\baselineskip}
\begin{figure}[t]
    \centering
    \includegraphics[width=\linewidth]{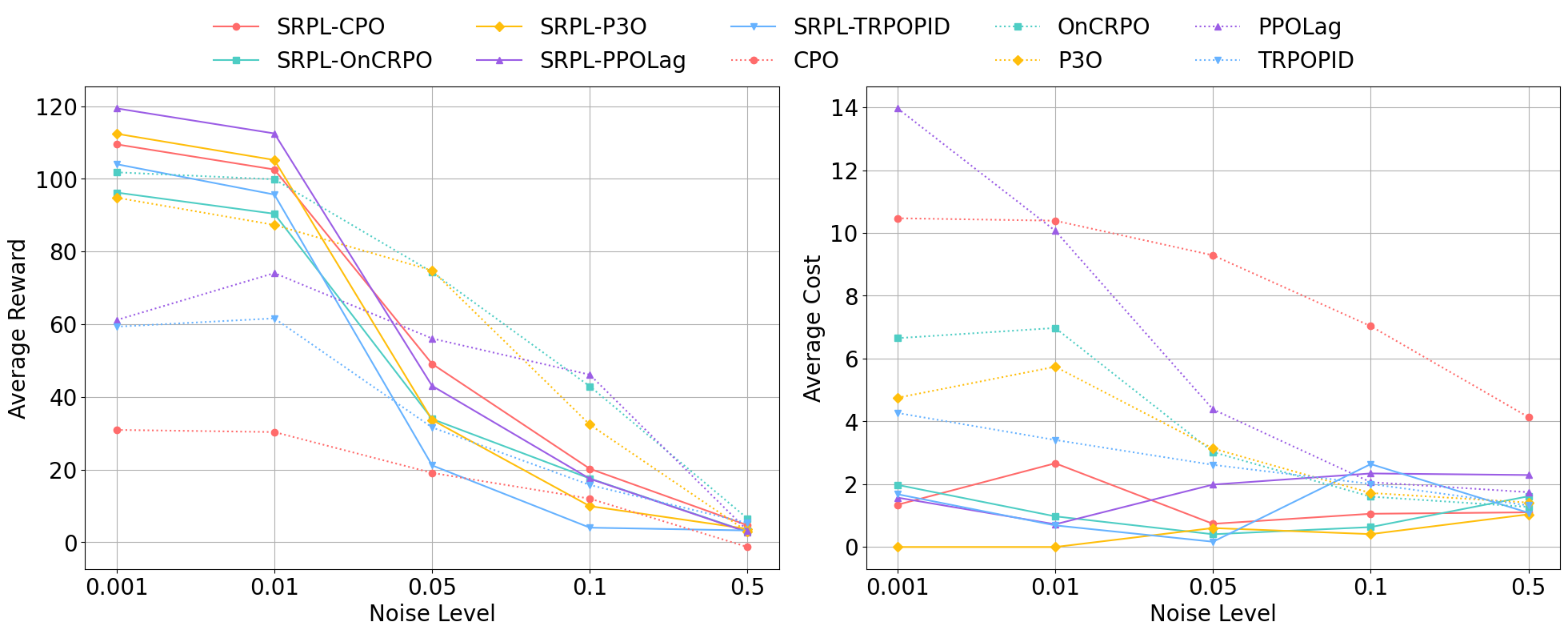}
    \caption{Average reward (left) and cost (right) versus Gaussian noise level ($\sigma$) applied to Lidar observations, evaluated on 500 WOMD scenarios per noise level. Higher rewards and lower costs indicate better performance.}
    \label{fig:robustness_comparison}
\end{figure}

\begin{table*}[t]
\caption{In-distribution evaluation results shown as mean (std) over 1,000 scenarios. \colorbox{blue!25}{Highlighted}: best performance per metric, \textbf{Bold}: SRPL improvements for cost and success rate (SR), *: statistically significant ($p < 0.05$, two-sided Wilcoxon signed-rank test, effect sizes reported in text).}
\label{tab:indistribution_results}
\centering
\begin{adjustbox}{max width=\textwidth}
\setlength{\tabcolsep}{5pt}
\renewcommand{\arraystretch}{1.2}
\begin{tabular}{@{}lc|c|ccccc|ccccc@{}}
\toprule
\textbf{Environment} & \textbf{Metric} & \textbf{Expert} & \multicolumn{5}{c|}{\textbf{Non-SRPL}} & \multicolumn{5}{c}{\textbf{SRPL}} \\
\cmidrule(lr){4-8} \cmidrule(lr){9-13}
& & & \textbf{CPO} & \textbf{PPOLag} & \textbf{TRPOPID} & \textbf{OnCRPO} & \textbf{P3O} & \textbf{CPO} & \textbf{PPOLag} & \textbf{TRPOPID} & \textbf{OnCRPO} & \textbf{P3O} \\
\midrule
\multirow{5}{*}{\textbf{WOMD}} 
& Reward ($\uparrow$) & 106.67 (95.23) & 76.30 (81.62) & 65.12 (76.10) & 51.88 (64.38) & 67.63 (75.94) & 83.76 (86.70) & 67.79 (71.62) & 81.01 (87.32) & 82.43 (84.33) & 65.92 (72.31) & \cellcolor{blue!25}88.72 (90.02) \\
& Cost ($\downarrow$)    & 0.16 (1.75)   & 2.00 (6.38)   & 2.03 (8.03)   & \cellcolor{blue!25}1.15 (3.97)   & 2.61 (9.13)   & 1.96 (6.65)   & \textbf{1.52 (4.58)*} & \textbf{1.54 (7.13)*} & 1.22 (4.52)   & \textbf{1.60 (7.89)*} & \textbf{1.25 (5.57)*} \\
& RC ($\uparrow$)   & 0.82 (0.35)   & \cellcolor{blue!25}0.74 (1.76)   & 0.71 (1.52)   & 0.59 (2.15)   & 0.60 (3.34)   & 0.67 (3.04)   & 0.60 (3.07)   & 0.62 (3.66)   & 0.72 (2.81)   & 0.60 (3.01)   & 0.68 (3.21)   \\
& SR ($\uparrow$)   & 1.00 (0.00)   & 0.88 (0.33)   & 0.81 (0.39)   & 0.67 (0.47)   & 0.82 (0.38)   & 0.89 (0.28)   & 0.82 (0.38)*  & \textbf{0.90 (0.30)}* & \textbf{0.92 (0.28)}* & \textbf{0.83 (0.39)}  & \cellcolor{blue!25}\textbf{0.93 (0.26)}  \\
& OOR ($\downarrow$)& 0.02 (0.12)   & 0.16 (0.37)   & 0.20 (0.40)   & 0.36 (0.48)   & 0.22 (0.41)   & 0.12 (0.33)   & 0.21 (0.41)   & 0.14 (0.34)   & 0.12 (0.33)   & 0.21 (0.41)   & \cellcolor{blue!25}0.09 (0.28)   \\
\midrule
\multirow{5}{*}{\textbf{NuPlan}} 
& Reward ($\uparrow$) & 88.70 (50.53)  & 56.42 (66.69) & 51.20 (47.46) & 41.56 (58.58) & \cellcolor{blue!25}72.12 (50.63) & 57.98 (65.56) & 52.95 (45.44) & 45.69 (43.64) & 48.00 (66.76) & 56.29 (56.06) & 62.56 (61.52) \\
& Cost ($\downarrow$)    & 0.07 (1.43)   & 3.26 (4.03)   & 1.57 (6.60)   & 2.49 (3.80)   & 1.56 (6.84)   & 3.24 (3.86)   & \textbf{1.33 (4.37)}  & \cellcolor{blue!25}\textbf{1.15 (4.13)*}  & 5.17 (5.04)*  & 3.52 (4.11)   & \textbf{2.75 (4.72)*} \\
& RC ($\uparrow$)   & 0.91 (0.23)   & \cellcolor{blue!25}0.96 (1.02)   & 0.74 (2.67)   & 0.53 (2.20)   & 0.95 (1.03)   & 0.86 (2.50)   & 0.90 (2.07)   & 0.81 (1.94)   & 0.63 (1.35)   & 0.93 (1.14)   & 0.72 (2.45)   \\
& SR ($\uparrow$)   & 1.00 (0.00)   & 0.79 (0.41)   & 0.72 (0.45)   & 0.52 (0.50)   & \cellcolor{blue!25}0.92 (0.27)   & 0.78 (0.27)   & 0.78 (0.36)  & 0.66 (0.47)*  & \textbf{0.71 (0.45)}* & 0.74 (0.44)*  & \textbf{0.85 (0.35)}* \\
& OOR ($\downarrow$)& 0.01 (0.08)   & 0.19 (0.39)   & 0.09 (0.29)   & 0.48 (0.50)   & 0.11 (0.31)   & 0.14 (0.34)   & 0.20 (0.68)   & 0.21 (0.41)   & 0.20 (0.40)   & 0.15 (0.36)   & \cellcolor{blue!25}0.08 (0.28)   \\
\bottomrule
\end{tabular}
\end{adjustbox}
\end{table*}

\subsubsection{Evaluation Metrics}
The performance of each agent is evaluated using five key metrics, averaged over 1,000 validation scenarios:
\begin{itemize}
\item \textbf{Reward:} The mean total reward accumulated per scenario, measuring overall driving task performance.
\item \textbf{Cost:} The mean total cost accumulated per scenario, reflecting the violations of safety constraints.
\item \textbf{Route Completion (RC):} The ratio of the agents' driven distance to the ground-truth trajectory length, averaged across scenarios \cite{li2023scenarionet}.
\item \textbf{Success Rate (SR):} The ratio of scenarios where the agent successfully reaches its destination \cite{li2023scenarionet}.
\item \textbf{Out of Road Rate (OOR):} The ratio of scenarios where the agent's vehicle drives off the designated drivable area.
\end{itemize}


Given the inherent stochasticity in deep reinforcement learning, comparing agents based solely on mean performance can be unreliable \cite{agarwal2021deep}. To ensure reliable evaluation, we conduct within-algorithm pairwise significance tests on Cost and Success Rate. For each SafeRL algorithm, baseline and SRPL-augmented agents are evaluated on the same 1,000 validation scenarios, yielding 1,000 paired observations per metric. We apply the two-sided Wilcoxon signed-rank test, appropriate for paired samples. The null hypothesis states that the median of paired differences is zero, the alternative hypothesis suggests a non-zero median, indicating performance change from SRPL augmentation. No multiple comparison correction is applied as each algorithm comparison addresses a distinct research question, focusing on within-algorithm patterns rather than universal superiority claims. Results with $p < 0.05$ are statistically significant. Effect sizes are calculated using $r = |z|/\sqrt{n}$, where $z$ is the standardized test statistic and $n$ is sample size. We interpret $|r| < 0.1$ as negligible, $0.1 \leq |r| < 0.3$ as small, $0.3 \leq |r| < 0.5$ as medium, and $|r| \geq 0.5$ as large effects following Cohen's conventions.

The implementation details and hyperparameters for the agents and the SRPL framework are as follows. A consistent training configuration was maintained across all algorithms to ensure a fair comparison. All agents were trained for 3.84 million environment steps over four random seeds. All agents' actor and critic networks consisted of hidden layers of [512, 256, 128]. We utilized normalized advantages and non-sequential scenarios during training to promote diverse exploration. For the SRPL framework specifically, the S2C model was configured with hidden layers of [256, 256, 128]. We adopted a safety horizon ($H_s = 60$) and a bin size of 2, based on the ablation analysis presented in \cite{mani2025safety}, which demonstrated that longer horizons and smaller bin sizes improve predictive accuracy for constraint violations in driving scenarios. Our implementation was built on top of the OmniSafe \cite{ji2024omnisafe}. Each training run was executed on a shared NVIDIA A100 GPU and took approximately 12-20 hours to complete. In addition, Expert metrics in Tab.~\ref{tab:indistribution_results} and Tab.~\ref{tab:crossdataset_results} represent performance from replaying recorded scenarios, serving as an upper bound for agents.

\begin{table*}[t]
\caption{Cross-dataset evaluation results (zero-shot transfer) shown as mean (std) over 1,000 scenarios. \colorbox{blue!25}{Highlighted}: best performance per metric, \textbf{Bold}: SRPL improvements for cost and success rate (SR), *: statistically significant ($p < 0.05$, two-sided Wilcoxon signed-rank test, effect sizes reported in text).}
\label{tab:crossdataset_results}
\centering
\begin{adjustbox}{max width=\textwidth}
\setlength{\tabcolsep}{5pt}
\renewcommand{\arraystretch}{1.2}
\begin{tabular}{@{}lc|c|ccccc|ccccc@{}}
\toprule
\textbf{Training \(\rightarrow\) Evaluation} & \textbf{Metric} & \textbf{Expert} & \multicolumn{5}{c|}{\textbf{Non-SRPL}} & \multicolumn{5}{c}{\textbf{SRPL}} \\
\cmidrule(lr){4-8} \cmidrule(lr){9-13}
& & & \textbf{CPO} & \textbf{PPOLag} & \textbf{TRPOPID} & \textbf{OnCRPO} & \textbf{P3O} & \textbf{CPO} & \textbf{PPOLag} & \textbf{TRPOPID} & \textbf{OnCRPO} & \textbf{P3O} \\
\midrule
\multirow{5}{*}{\textbf{NuPlan \(\rightarrow\) WOMD}} 
& Reward ($\uparrow$) & 106.67 (95.23) & 61.21 (75.19) & 56.88 (74.75) & 33.29 (57.59) & 72.71 (96.63) & 65.40 (79.33) & 49.64 (63.95) & 39.21 (55.56) & 44.61 (76.78) & 55.90 (73.57) & \cellcolor{blue!25}75.62 (78.07) \\
& Cost ($\downarrow$) & 0.16 (1.75) & 2.12 (6.64) & 3.49 (12.23) & 2.95 (9.63) & 7.61 (19.98) & 2.22 (8.69) & 2.63 (8.11) & \cellcolor{blue!25}\textbf{1.45 (4.20)*} & 5.43 (16.31) & \textbf{3.81 (12.73)*} & 2.68 (12.01) \\
& RC ($\uparrow$) & 0.82 (0.35) & 0.75 (3.49) & 0.73 (1.93) & 0.21 (3.00) & 0.67 (4.19) & 0.69 (2.64) & 0.55 (3.00) & 0.37 (2.85) & 0.62 (2.79) & 0.54 (3.70) & \cellcolor{blue!25}0.82 (2.36) \\
& SR ($\uparrow$) & 1.00 (0.00) & 0.76 (0.42) & 0.75 (0.43) & 0.42 (0.49) & 0.90 (0.30) & 0.74 (0.44) & 0.72 (0.45)* & 0.60 (0.49)* & \textbf{0.69 (0.46)}* & 0.71 (0.45)* & \cellcolor{blue!25}\textbf{0.92 (0.27)}* \\
& OOR ($\downarrow$) & 0.02 (0.12) & 0.27 (0.45) & 0.24 (0.43) & 0.61 (0.49) & 0.16 (0.36) & 0.30 (0.46) & 0.31 (0.46) & 0.40 (0.49) & 0.35 (0.48) & 0.26 (0.44) & \cellcolor{blue!25}0.10 (0.30) \\
\midrule
\multirow{5}{*}{\textbf{WOMD \(\rightarrow\) NuPlan}} 
& Reward ($\uparrow$) & 88.70 (50.53) & 60.55 (49.35) & 51.14 (61.64) & 31.85 (94.87) & 55.54 (57.98) & 58.96 (65.84) & \cellcolor{blue!25}61.68 (45.55) & 31.22 (39.27) & 46.82 (62.03) & 54.34 (71.58) & 39.42 (42.13) \\
& Cost ($\downarrow$) & 0.07 (1.43) & 2.13 (3.85) & 2.85 (6.17) & 9.70 (6.35) & 2.50 (9.62) & 3.23 (6.89) & \cellcolor{blue!25}\textbf{0.86 (2.95)*} & \textbf{1.61 (6.71)*} & \textbf{3.25 (5.22)*} & 4.56 (6.97)* & \textbf{1.79 (9.22)} \\
& RC ($\uparrow$) & 0.91 (0.23) & 0.85 (2.66) & 0.74 (1.04) & 0.57 (2.14) & 0.64 (2.72) & 0.89 (1.03) & 0.91 (3.57) & 0.52 (1.68) & 0.59 (2.48) & \cellcolor{blue!25}0.97 (2.07) & 0.50 (3.01) \\
& SR ($\uparrow$) & 1.00 (0.00) & 0.79 (0.41) & 0.65 (0.48) & 0.56 (0.50) & 0.75 (0.43) & 0.75 (0.43) & \cellcolor{blue!25}\textbf{0.83 (0.37)*} & 0.49 (0.50)* & \textbf{0.65 (0.48)}* & \textbf{0.77 (0.42)} & 0.56 (0.50)* \\
& OOR ($\downarrow$) & 0.01 (0.08) & 0.10 (0.30) & 0.10 (0.29) & 0.27 (0.44) & 0.17 (0.37) & 0.10 (0.30) & 0.14 (0.35) & \cellcolor{blue!25}0.04 (0.18) & 0.10 (0.29) & 0.11 (0.31) & 0.05 (0.22) \\
\bottomrule
\end{tabular}
\end{adjustbox}
\end{table*}

\subsection{In-Distribution Evaluation}

The learning dynamics on WOMD, shown in Fig.~\ref{fig:baseline_combined} and Fig.~\ref{fig:srpl_combined}, reveal that SRPL-augmented agents exhibit steeper initial learning curves compared to their baselines, indicating that predictive safety information provides an inductive bias resulting in accelerated learning and improved sample efficiency. The evaluation results on WOMD, detailed in Tab.~\ref{tab:indistribution_results}, demonstrate that SRPL can improve SafeRL performance across multiple algorithms. SRPL yields statistically significant improvements in the success rate for PPOLag $(r = 0.85, p < 0.05)$ and TRPOPID $(r = 0.86, p < 0.05)$. SRPL-TRPOPID shows better reward-safety trade-off with a 58.9\% increase in reward at the expense of a 6\% cost increase. SRPL-PPOLag is the only algorithm to demonstrate statistically significant improvements in both success rate and cost $(r = 0.78, p < 0.05)$.

The evaluation results on NuPlan, detailed in Tab.~\ref{tab:indistribution_results}, reveal different performance patterns, demonstrating that the effectiveness of SRPL is sensitive to the dataset characteristics. The algorithmic hierarchy observed on WOMD does not hold on NuPlan. For instance, while SRPL-PPOLag achieves a statistically significant reduction in cost on NuPlan $(r = 0.70, p < 0.05)$, this improvement comes at the expense of reduced success rate, contrasting with its improvements on WOMD. The most consistent positive outcome on NuPlan is observed with SRPL-P3O, which achieves statistically significant improvements in both, the success rate $(r = 0.85, p < 0.05)$ and the cost $(r = 0.83, p < 0.05)$.

Notably, SRPL-CPO exhibits consistently reduced success rate across both datasets when compared to its baseline CPO, presenting a distinct pattern from other algorithms. On WOMD, while SRPL improves success rates for the other four algorithms, SRPL-CPO shows reduced reward and success rate, though with improved cost performance, indicating that CPO augmented with SRPL becomes conservative. This pattern persists on NuPlan, where SRPL-CPO again demonstrates reduced reward and success rate while maintaining better cost control. This systematic underperformance in complex real-world driving scenarios contrasts with SRPL-CPO's effectiveness in controlled benchmark environments \cite{mani2025safety}, suggesting that the combination of CPO's trust-region constraints with SRPL's predictive representations may create conservatism specifically in high-dimensional environments where constraint interactions are more complex.

These results reveal algorithm-specific interactions with SRPL: while P3O demonstrates the most consistent improvements across datasets, achieving cost reductions and success rate improvements on both WOMD and NuPlan, CPO exhibits conservative behavior that reduces task performance despite safety gains.

\subsection{Zero-Shot Cross-Dataset Evaluation}
To assess generalization to unseen driving domains, we performed zero-shot transfer evaluation where agents trained on one dataset were evaluated on another without fine-tuning. Results are presented in Tab.~\ref{tab:crossdataset_results}.

A primary observation is the reduced performance that occurs when agents are transferred to an unseen domain. This effect, however, is asymmetric. When NuPlan-trained agents are evaluated in the more diverse WOMD, they exhibit a notable decline in reward and success rates, alongside an increase in out-of-road instances relative to their WOMD-trained counterparts.

Conversely, the transfer from WOMD-trained agents to NuPlan, while also showing a general reduction in reward, indicates certain areas of positive transfer. In particular, agents trained on WOMD exhibit a reduced out-of-road rate when evaluated on NuPlan compared to agents trained directly on NuPlan. Additionally, both CPO and SRPL-CPO agents trained on WOMD outperform their counterparts trained directly on NuPlan when evaluated on the NuPlan dataset, with SRPL-CPO demonstrating statistically significant improvements in both cost reduction $(r = 0.83, p < 0.05)$ and success rate $(r = 0.84, p < 0.05)$. Similarly, SRPL-OnCRPO trained on WOMD achieves higher route completion on NuPlan than the same algorithm trained directly on NuPlan.

This asymmetric transfer performance is consistent with the hypothesis that more diverse training data leads to better generalization. Supporting this hypothesis, WOMD scenarios exhibit greater environmental diversity: an average of 85.18 vehicles per scenario compared to 52.18 in NuPlan, longer average track lengths (130.45 meters relative to 90.42 meters), and higher intersection density (0.69 compared to 0.54).

Despite the challenges of domain shift, augmenting agents with SRPL demonstrates improved generalization capabilities compared to non-SRPL methods in several cases. Across both transfer directions, SRPL-augmented agents show improved performance on multiple evaluation metrics, though no single algorithm achieves optimal performance across all metrics. These findings suggest that predictive safety representations can be an effective strategy for mitigating adverse effects of domain shift, although the benefits vary by algorithm and evaluation metric.

\setlength{\abovecaptionskip}{\baselineskip}
\setlength{\belowcaptionskip}{\baselineskip}
\begin{figure*}[t]
\centering

\setlength{\abovecaptionskip}{5pt}
\setlength{\belowcaptionskip}{0pt}

\begin{subfigure}{0.49\textwidth}
    \includegraphics[width=\linewidth]{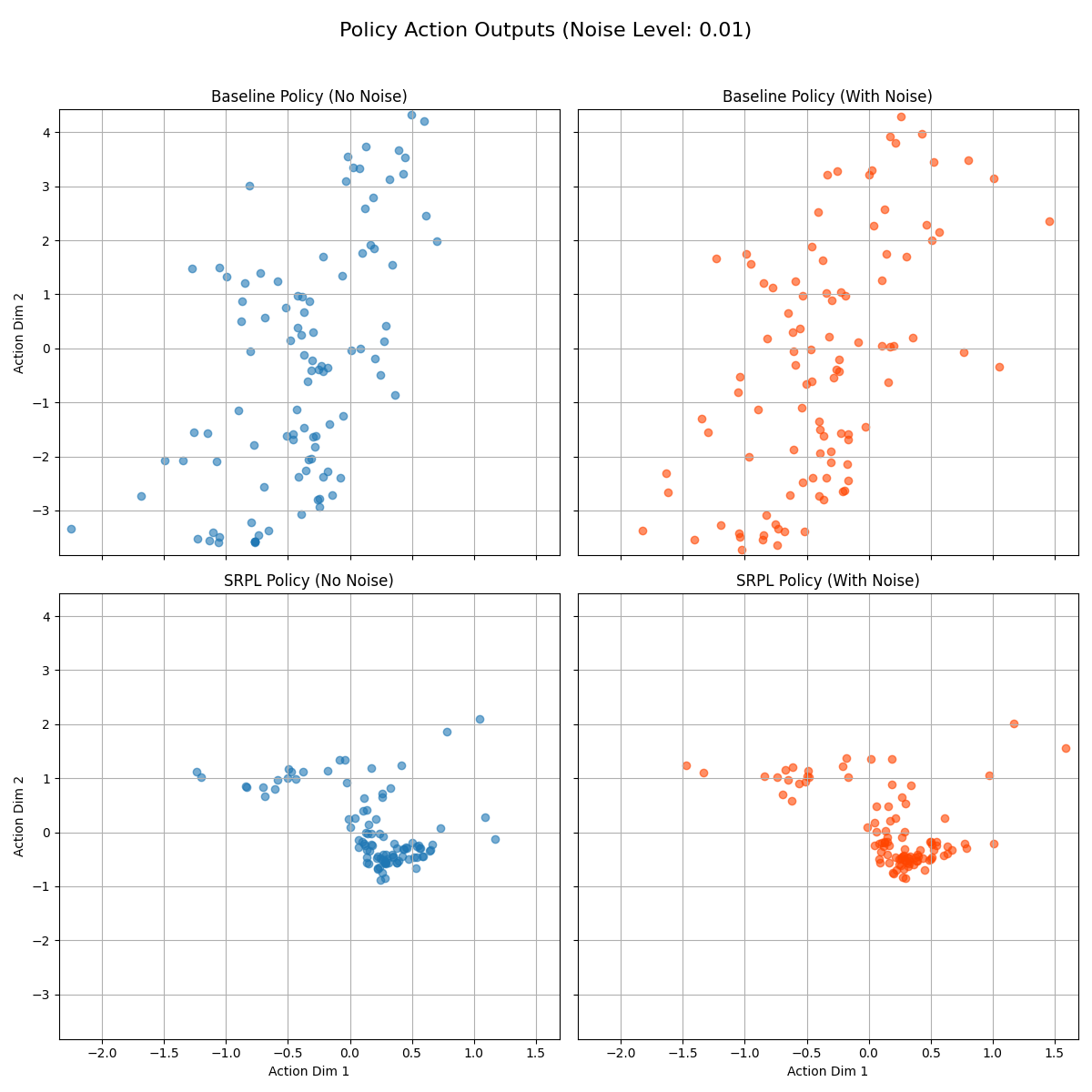}
    \caption{Policy action outputs at noise level $\sigma = 0.01$ comparing baseline CPO and SRPL-CPO under clean and noisy conditions.}
\end{subfigure}
\hfill 
\begin{subfigure}{0.49\textwidth}
    \includegraphics[width=\linewidth]{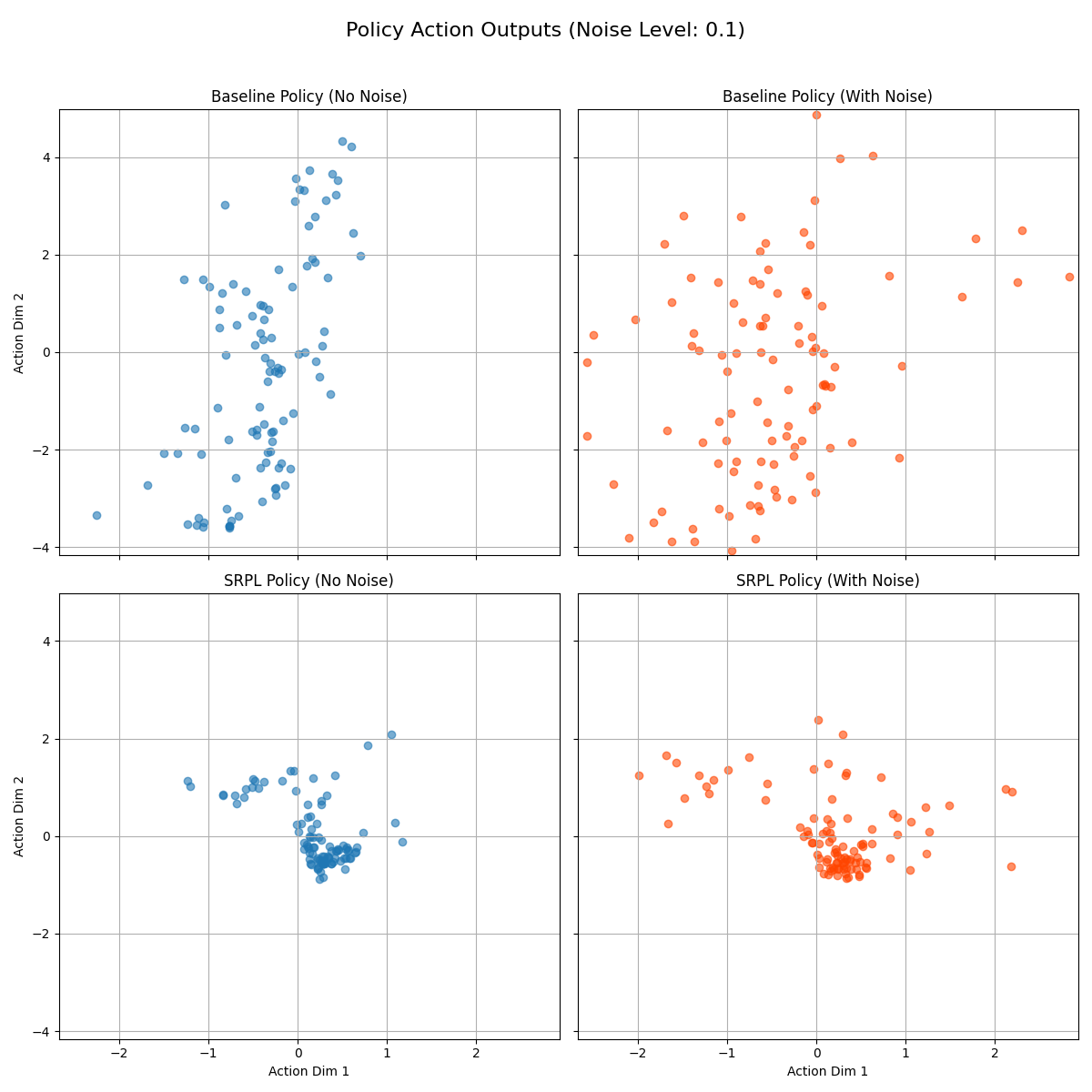}
    \caption{Policy action outputs at noise level $\sigma = 0.1$ comparing baseline CPO and SRPL-CPO under clean and noisy conditions.}
\end{subfigure}

\caption{Comparison of baseline and SRPL-augmented mean action outputs for different noise levels.}
\label{fig:policy_outputs}
\end{figure*}

\subsection{Robustness to Observational Noise}
To be deployed in the real-world, an autonomous driving agent must demonstrate robustness to sensor noise. Real-world conditions, such as rain or fog, can distort Lidar point clouds, while sudden lighting changes, like exiting a tunnel, or lens flare at night, can compromise camera images. These conditions introduce noise into the agent's input, leading it to make unsafe decisions.

To evaluate agent performance under noisy conditions, we conducted a robustness experiment where the Lidar input was systematically perturbed. Specifically, Gaussian noise with a varying noise level, $\sigma$, was injected into the observation. For each noise level, agents trained on WOMD were evaluated on 500 validation WOMD scenarios.

The results, presented in Fig.~\ref{fig:robustness_comparison}, indicate that SRPL improves robustness to observation noise. The SRPL-augmented variants maintain higher rewards and lower costs at low to moderate noise levels. The cost plot, in particular, shows that the SRPL-augmented variants maintain low costs (in the 0-3 range) as the noise increases, while the baselines show an increase in constraint violations. This robustness is an important step towards practical viability, indicating that SRPL can improve performance in a clean environment and make agents more resilient to the observation noise inherent to the physical systems. To understand this resilience better, we analyze the stabilizing effect of SRPL on policy outputs in the following subsection.

\subsection{Analysis of Policy Output Stabilization}
To understand the mechanism underlying SRPL's robustness to noise, we investigate whether predictive safety information acts as a stabilizing signal. SRPL augments observations with learned safety predictions, providing additional information that may reduce sensitivity to noise perturbations. Our robustness analysis reveals that baseline CPO's cost increases significantly with added noise while SRPL-CPO remains nearly stable compared to the noise-free environment. In Fig.~\ref{fig:robustness_comparison}, CPO's cost rises from approximately 2.0 in noise-free conditions to over 10.0 even at relatively low noise levels, while SRPL-CPO maintains costs below 3.0 across all tested noise levels.

To quantify this stabilization effect, we analyze mean action outputs at two noise levels: $\sigma = 0.01$ and $\sigma = 0.1$, representing 1\% and 10\% of the observation range [0, 1]. Action outputs were collected from 100 observations across different driving scenarios for each noise level. As shown in Fig.~\ref{fig:policy_outputs}, baseline CPO exhibits high output variance when comparing clean versus noisy inputs, with actions distributed across a wider action space region. In contrast, SRPL-CPO maintains lower variance in its action outputs under noise, with SRPL reducing variance by 3.5 percentage points compared to baseline at $\sigma = 0.01$ and by 79.7 percentage points at $\sigma = 0.1$, indicating implicit stabilization that reduces sensitivity to observation perturbations. However, this stabilization effect has limits, as performance diminish at sufficiently high noise levels where noise begins to overwhelm the input signal.

These findings suggest that SRPL's predictive safety representations enable more consistent policy behavior under moderate noise conditions, providing insight into the mechanism underlying the robustness improvements observed in Fig.~\ref{fig:robustness_comparison}.

\section{CONCLUSION}
This work presented a systematic evaluation of augmenting on-policy SafeRL agents with predictive safety representations using the SRPL framework for autonomous driving. Our experiments on the WOMD and NuPlan datasets revealed three primary findings. First, SRPL improved the reward-safety tradeoff, particularly for success rate improvements, with statistically significant effect sizes ranging from $(r = 0.78-0.86, p < 0.05)$ for in-distribution evaluation and $(r = 0.65-0.84, p < 0.05)$ for cross-dataset evaluation. However, this effectiveness proved to be context-dependent, varying with the underlying policy optimization method and environmental characteristics. Our second finding demonstrated that data diversity influences generalization, as agents trained on the more diverse WOMD dataset transferred more effectively to NuPlan than the reverse. Third, SRPL improved robustness against Gaussian noise by providing implicit stabilization that reduced policy output variance.

Building on these findings, we present practical guidance for algorithm selection in SafeRL for autonomous driving. When cost reduction is the primary objective, SRPL-PPOLag consistently achieves improved cost performance across in-distribution, cross-dataset, and robustness evaluations compared to its baseline, achieving cost reductions with effect sizes $(r = 0.70-0.83, p < 0.05)$, making it the recommended choice for safety-critical deployments. For scenarios prioritizing success rates, OnCRPO without SRPL achieves the highest or near-highest success rates across both in-distribution and cross-dataset evaluations. In the WOMD environment, SRPL-P3O emerges as the top performer, suggesting its suitability for diverse scenarios. For deployment requiring robustness to observation noise, specifically Gaussian noise, SRPL augmentation should be preferred.

These findings contribute to understanding how SRPL functions in complex driving scenarios, though the scope of this study presents certain limitations. The experiments were conducted on a fixed number of scenarios, and future work requires investigation of scalability to larger-scale datasets. The robustness analysis was limited to Gaussian noise, and performance evaluation against more realistic sensor corruptions remains necessary. Finally, the algorithm and environment-specific nature of SRPL's effectiveness highlights a key challenge for future research: the development of universal safety representations that demonstrate effectiveness across a broader range of algorithms and driving domains.

\bibliographystyle{IEEEtran}
\bibliography{updated_references}

\end{document}